\documentclass[runningheads]{llncs}
\usepackage[T1]{fontenc}
\usepackage{graphicx}
\usepackage{setspace}  % \begin{spacing}
\usepackage{booktabs}  % \toprule \midrule \bottomrule
\usepackage{adjustbox}  % adjustbox
\usepackage{multirow}  % multirow
\usepackage{amsmath}  % boldsymbol
\usepackage{amssymb}  % \triangleq
\usepackage{xcolor}  % \textcolor
\usepackage{cite}  % \cite{a, b, c, ...}
\usepackage{relsize}  % \mathlarger

\begin{document}
\title{CoT-BERT: Enhancing Unsupervised Sentence Representation through Chain-of-Thought}
\author{Bowen Zhang \and Kehua Chang \and Chunping Li}
\authorrunning{Bowen et al.}  % Part of LEFT running header
\titlerunning{Enhancing Unsupervised Sentence Representation through CoT}  % Part of RIGHT running header
\institute{
School of Software, Tsinghua University\\
\email{zbw23@mails.tsinghua.edu.cn}\\
\email{changkehua@gmail.com}\\
\email{cli@tsinghua.edu.cn}
}
\maketitle              % typeset the header of the contribution
\begin{abstract}

Unsupervised sentence representation learning aims to transform input sentences into fixed-length vectors enriched with intricate semantic information while obviating the reliance on labeled data. Recent strides within this domain have been significantly propelled by breakthroughs in contrastive learning and prompt engineering. Despite these advancements, the field has reached a plateau, leading some researchers to incorporate external components to enhance the quality of sentence embeddings. Such integration, though beneficial, complicates solutions and inflates demands for computational resources. In response to these challenges, this paper presents CoT-BERT, an innovative method that harnesses the progressive thinking of Chain-of-Thought reasoning to tap into the latent potential of pre-trained models like BERT. Additionally, we develop an advanced contrastive learning loss function and propose a novel template denoising strategy. Rigorous experimentation demonstrates that CoT-BERT surpasses a range of well-established baselines by relying exclusively on the intrinsic strengths of pre-trained models. \footnote{Our code and checkpoints are available at \url{https://github.com/ZBWpro/CoT-BERT}.}

\end{abstract}
\section{Introduction}

Sentence representation is a fundamental aspect of Natural Language Processing (NLP), which concentrates on converting input sentences into fixed-length numerical vectors, commonly known as sentence embeddings. These vectors are crucial for enabling computational systems and advanced deep learning models to process linguistic data effectively. Their applications span across various domains, including information retrieval, text clustering, topic modeling, recommendation systems, and the development of artificial intelligence agents \cite{Retrieval-based-LM-ACL-2023}. Moreover, they serve as essential features for neural networks engaged in downstream tasks such as relationship extraction, sentiment classification, and text implication recognition.

In this research field, unsupervised sentence representation learning has garnered substantial attention owing to its independence from annotated data. Currently, the prevailing paradigm in academia involves leveraging pre-trained language models (PLMs) like BERT \cite{BERT-NAACL-2019} and RoBERTa \cite{RoBERTa-2019}, with contrastive learning to mitigate the issue of semantic space anisotropy \cite{BERT-flow-EMNLP-2020} by drawing similar samples closer while pushing dissimilar ones apart.

With the rise of prompt engineering, a series of studies have been initiated to integrate prompts into sentence representation tasks. A remarkable endeavor, PromptBERT \cite{PromptBERT-EMNLP-2022}, adopts a manual template ``This sentence: `[X]' means [MASK].'' to encapsulate the input sentence [X], using the output vector corresponding to the [MASK] token as the original sentence's representation. Despite its impressive performance, subsequent advancements have encountered a plateau. For instance, ConPVP \cite{ConPVP-EMNLP-2022}, an improvement on PromptBERT, achieves a marginal gain, with only a 0.03 increment in the average Spearman's correlation score across seven Semantic Textual Similarity (STS) benchmarks when implemented with RoBERTa$_{\rm base}$.

In pursuit of further enhancements, some research \cite{DiffCSE-NAACL-2022, RankCSE-ACL-2023, RankEncoder-ACL-2023, DenoSent-AAAI-2024} has explored incorporating external models or corpora to assist in sentence embedding derivation. Although these additions can improve model performance, they concurrently increase solution complexity and computational resource consumption. For example, the current state-of-the-art (SOTA) work, RankCSE \cite{RankCSE-ACL-2023}, necessitates knowledge distillation from two high-capacity teacher models for training. Specifically, the development of RankCSE-BERT$_{\rm base}$ involves training both SimCSE-BERT$_{\rm base}$ \cite{SimCSE-EMNLP-2021} and SimCSE-BERT$_{\rm large}$, constraining its applicability due to the computational demands of updating BERT$_{\rm large}$. Another cutting-edge strategy, RankEncoder \cite{RankEncoder-ACL-2023}, requires the introduction of a large external corpus on top of the training set, which poses challenges in data-scarce scenarios.

This paper endeavors to fully unleash the potential of pre-trained models, achieving results on par with or superior to previous methods without introducing any external components. To this end, we draw inspiration from the progressive thinking of Chain-of-Thought (CoT) \cite{CoT-NIPS-2022}, attempting to derive sentence representations through a multi-staged process.

CoT, an emerging prompting technique, has shown promise in augmenting the performance of high-parameter generative models in complex reasoning tasks by breaking down problems into a series of logical steps leading to the final answer. Although BERT operates with significantly fewer parameters compared to models such as LLaMA \cite{Llama-2023} and GPT \cite{GPT-3-NIPS-2020}, we posit that the progressive nature of CoT can be adapted to discriminative models. Given BERT's capability to assign context-specific meanings to the special token [MASK] via its attention mechanism during the masked language modeling (MLM) task, this token exhibits an adaptive characteristic that aligns well with the CoT methodology.

Accordingly, we propose a two-step approach to sentence representation, which includes both comprehension and summarization stages, with the latter's output serving as the representation for the original sentence. Extensive experimental evaluations indicate that this method, termed CoT-BERT, outperforms several robust baselines without necessitating additional parameters. The primary contributions of this paper are outlined as follows:
\begin{itemize}
\item \textbf{Introduction of CoT-BERT}: This study unveils CoT-BERT, a groundbreaking method that integrates the progressive logic of Chain-of-Thought with sentence representation. Our approach demonstrates performance that matches or exceeds current rank-based SOTA solutions, while obviating the need for external corpora or auxiliary text representation models. Notably, employing RoBERTa$_{\rm base}$ as the PLM, we attain a Spearman correlation of 80.62\% across seven STS tasks, markedly surpassing the existing best result. To our knowledge, CoT-BERT represents the inaugural effort to amalgamate CoT reasoning with sentence representation.

\item \textbf{Extended InfoNCE Loss}: We present a superior contrastive learning loss function that extends beyond the conventional InfoNCE Loss \cite{InfoNCELoss-2018}. This function facilitates an enhanced optimization of the PLM’s semantic space uniformity by introducing contrast not only between anchor sentences and negative samples but also between positive and negative instances.

\item \textbf{Innovative Template Denoising Strategy}: Our research proposes an advanced template denoising strategy to eliminate the potential influence of semantic interpretation attributable to prompt biases. This objective is realized by filling blank templates with [PAD] placeholders of identical length to the input sentence and adjusting the attention masks accordingly.

\item \textbf{Comprehensive Experimental Evaluation}: We conduct a thorough examination of CoT-BERT's performance across seven established STS benchmarks. Additionally, we provide exhaustive ablation studies and analyses focused on the main innovations of CoT-BERT to ascertain the reasons behind its effectiveness. Furthermore, our code and checkpoints have been made available for replication and further experimentation.
\end{itemize}

\section{Related Work}

This section reviews three categories of research directly related to our work: unsupervised contrastive learning methods for sentence embeddings, techniques for text representation through prompts, and chain-of-thought prompting for multi-step reasoning.

\subsection{Unsupervised Sentence Representation}

Inspired by advancements in computer vision, numerous studies have explored the application of contrastive learning to unsupervised sentence representation \cite{ArcCSE-ACL-2022, ESimCSE-COLING-2022, DCLR-ACL-2022, PCL-EMNLP-2022}. Compared to post-processing strategies like BERT-flow \cite{BERT-flow-EMNLP-2020} and BERT-whitening \cite{BERT-whitening-2021}, contrastive learning demonstrates more pronounced improvements within the semantic space of BERT, thus establishing itself as the predominant method for deriving embeddings in the current NLP community. Among these efforts, ConSERT \cite{ConSERT-ACL-2021} employs four strategies, including adversarial attacks and token shuffling, to construct positive samples. SimCSE \cite{SimCSE-EMNLP-2021}, on the other hand, utilizes standard dropout for minimal data augmentation. Building upon this, SSCL \cite{SSCL-ACL-2023} leverages outputs from the intermediate layers of PLMs as hard negatives to mitigate over-smoothing. In general, these methods primarily innovate in the construction of positive and negative samples, yet they do not fully exploit BERT's behavior during its pre-training phase.

A new frontier in this field involves integrating external components to refine sentence embeddings. DiffCSE \cite{DiffCSE-NAACL-2022}, for instance, introduces a generator and discriminator structure atop the encoder, aiming to endow the model with the ability to distinguish between original and edited sentences. RankCSE \cite{RankCSE-ACL-2023} proposes utilizing multiple text representation models for knowledge distillation, coupled with a training methodology that combines ranking consistency and contrastive learning. Moreover, RankEncoder \cite{RankEncoder-ACL-2023} suggests the inclusion of an external corpus to enrich sentence representation derivation. While methods that incorporate external models or datasets typically exhibit enhanced performance over those relying solely on PLMs, this integration inevitably increases the complexity and resource demands of their frameworks.

\subsection{Prompt-based Learning}

Originating from GPT models \cite{GPT-3-NIPS-2020}, the concept of prompts has rapidly expanded into diverse domains, including semantic textual similarity. Prompt-based techniques are designed to maximize the utilization of prior knowledge stored within PLMs. For models like BERT and RoBERTa, this objective is chiefly realized through the transformation of downstream tasks into formats that closely emulate MLM. Notably, PromptBERT \cite{PromptBERT-EMNLP-2022} employs a manually designed template to encapsulate the input sentence and takes the output vector corresponding to the [MASK] token as the representation. In a parallel endeavor, PromCSE \cite{PromCSE-EMNLP-2022} deploys a distinctive approach by freezing the entire pre-trained model while integrating multi-layer learnable soft prompts. Furthermore, ConPVP \cite{ConPVP-EMNLP-2022} merges continuous prompts with various manual templates. These elements are then converted into anchor sentences, positive instances, and negative instances for incorporation into the InfoNCE Loss. 

\subsection{Chain-of-Thought Prompting}

Chain-of-Thought (CoT) prompting represents a revolutionary approach, guiding large language models (LLMs) through a series of intermediate steps towards the final answer. As CoT is primarily targeted at direct inference scenarios and typically exhibits significant advantages only when the model reaches a substantial size, CoT reasoning is considered an emergent property of LLMs \cite{CoT-NIPS-2022}.

Despite this, the principle of decomposing complex problem, as advocated by CoT, finds broad applicability in deep learning. As an illustration, consider the representation of text in neural networks, wherein the journey commences with the encoding of individual lexical units before advancing toward holistic sentence representation. Therefore, we aspire to extend this technique to discriminative models like BERT to further unlock the potential inherent within PLMs.

\section{Methodology}
\label{sec:methodology}

In this section, we initially present our CoT-style two-stage sentence representation strategy, along with its underlying design principles, in subsection~\ref{sec:two_stage_representation}. Following this, in subsection~\ref{sec:extended_infonce_loss}, we proceed to introduce the extended InfoNCE Loss, accompanied by an intuitive explanation for its heightened performance. Eventually, our refined template denoising technique is expounded upon in subsection~\ref{sec:template_denoising}.

\subsection{Chain-of-Thought-Styled Two-stage Sentence Representation}
\label{sec:two_stage_representation}

To effectively adapt CoT to new application scenarios, it is essential to clarify its main characteristics. Formally, CoT consists of two principal components: the reasoning process and the resultant conclusions. The former is a model-generated sequence predicated on given context, while the latter constitutes the final outcome derived from synthesizing the intermediate reasoning steps. Collectively, they form a problem-solving strategy that progresses from simple to complex.

CoT primarily operates on high-parameter generative PLMs. Although discriminative models like BERT possess robust natural language understanding capabilities, they cannot directly generate intermediate reasoning steps. Therefore, to guide such models towards CoT-style progressive reasoning, we need to employ prompt engineering to artificially design a multi-step sentence representation pipeline that is both versatile and adaptive.

Versatility here refers to the method's applicability to various sentences types, which requires a general strategy for solving sentence representation tasks. Adaptability, in contrast, emphasizes the model's ability to identify and prioritize key information from sentences based on their distinct semantics. Drawing inspiration from human practices in text summarization, we propose a two-stage sentence embedding derivation methodology: comprehension followed by summarization. Each stage incorporates a [MASK] token, enabling the model to attribute contextually relevant interpretations to [MASK] across diverse scenarios.

Following the aforementioned principles, our devised templates are detailed in Table~\ref{tab:templates}. Given that the training set for unsupervised sentence representation tasks only includes a series of standalone sentences, different templates are adopted to construct anchor sentences, positive instances, and hard negative instances. These elements are subsequently integrated into our revised contrastive learning loss function for training, a procedure elaborated upon in subsection~\ref{sec:extended_infonce_loss}. After forward computation, we take the output vector corresponding to the final [MASK] token as the sentence embedding for the input text, mirroring the essence of CoT reasoning, which focuses on the model's conclusive output rather than the intermediate process.
\begin{table}[htbp]
\caption{Manual templates for our prompt-based learning. To alleviate burden and ensure conciseness in CoT-BERT, the templates selected for the anchor sentence, positive instance, and hard negative instance only have slight differences.}
\centering
\begin{spacing}{1.5}
\begin{adjustbox}{width=1.0\linewidth, center}
    \begin{tabular}{c}
    \toprule
    {\bf Anchor Sentence}  \\
    The sentence \textcolor{red}{of} ``[X]'' means [MASK], so it can be summarized as [MASK]. \\
    \midrule 
    {\bf Positive Instance} \\
    The sentence \textcolor{red}{:} ``[X]'' means [MASK], so it can be summarized as [MASK]. \\
    \midrule 
    {\bf Hard Negative Instance} \\
    The sentence : ``[X]'' \textcolor{red}{does not} mean [MASK], so it \textcolor{red}{cannot} be summarized as [MASK]. \\
    \bottomrule 
    \end{tabular}
\end{adjustbox}
\end{spacing}
\label{tab:templates}
\end{table}

In section~\ref{sec:experiment}, this paper will further substantiate the rationality of CoT-BERT's two-stage sentence representation method and its correlation with CoT through an array of experiments. Specifically, we aim to explore the following pivotal inquiries:
\begin{itemize}
\item Does the two-stage sentence representation strategy of CoT-BERT surpass the use of any single sub-stage alone?
\item Does the progressive relationship between sub-stages contribute to enhanced model performance? If affirmative, can a genuine progressive linkage between comprehension and summarization be established?
\item Compared to static templates, does the adaptive reasoning facilitated by the [MASK] token exhibit better universality?
\end{itemize}

Exploring these questions will deepen our understanding of CoT-inspired multi-stage reasoning. Empirical validation is critical for providing definitive answers to these considerations.

\subsection{Enhancing Contrast with Extended InfoNCE Loss}
\label{sec:extended_infonce_loss}

Current methods for unsupervised sentence representation through contrastive learning commonly employ the InfoNCE Loss to guide the training process. For any given sentence $x_i$ within a batch of size $N$, where $f(\cdot)$ denotes the encoding technique and $\tau$ represents a temperature coefficient, the InfoNCE Loss is defined as follows:
\begin{equation}
\label{eq:info_nce}
     \mathcal{\ell}_{i} = - {\log} \frac{e^{{\text{sim}}(f(x_i),f(x_i)^+)/\tau}}{\sum_{j=1}^Ne^{\text{sim}(f(x_i),f(x_j)^+)/\tau}}
\end{equation}

In Equation~\ref{eq:info_nce}, $\text{sim}$ signifies a similarity metric between two sentence vectors, typically cosine similarity. This formula encourages the model to maximize the similarity between the anchor sentence vector $f(x_i)$ and its positive instance $f(x_i)^+$, while simultaneously diminishing the similarity between $f(x_i)$ and other unrelated sentence vectors $f(x_j)$ within the same batch. Both SimCSE and PromptBERT adopt an InfoNCE Loss of this form. In SimCSE, positive instances are crafted using the intrinsic dropout of Transformer blocks, whereas PromptBERT employs different templates to generate positive samples. Although PromptBERT's comparative experiments reveal minimal performance distinctions between these two methods under their manual templates, the utilization of varied templates for data augmentation can additionally facilitate the creation of hard negatives.

\begin{figure*}[htbp]
\centering \includegraphics[width=1.0\linewidth]{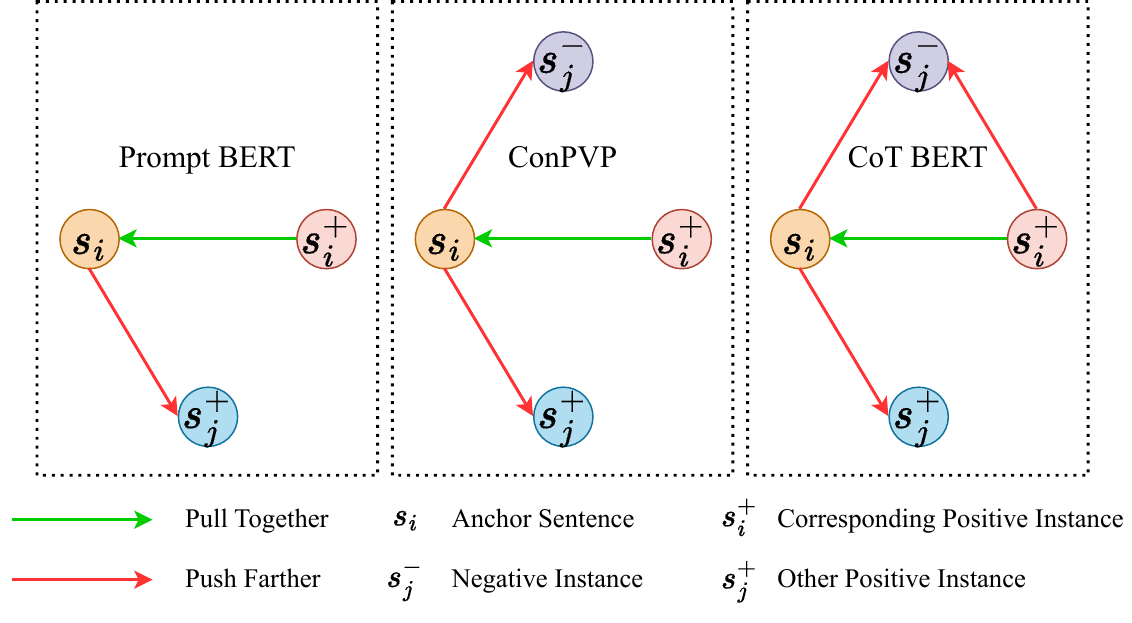}
\caption{Behavior of three variants of InfoNCE Loss within the semantic space of BERT. For clarity, we depict this figure with the anchor sentence $s_i$ as the focal point.}
\label{fig:loss_functions}
\end{figure*}

Specifically, ConPVP designs multiple templates and integrates the comparison between the anchor sentence vector $f(x_i)$ and the negative instance $f(x_j)^-$ in the denominator of Equation~\ref{eq:info_nce}. By contrast, our CoT-BERT extends this by also introducing a comparison between the positive instance $f(x_i)^+$ and the negative sample $f(x_j)^-$. The differences among these approaches are visualized in Figure~\ref{fig:loss_functions}. Intuitively, our refined InfoNCE Loss incorporates more reference items into the sentence representation calculation process, rendering the distribution disparities among unrelated sentence vectors more pronounced within the semantic space. Subsection~\ref{sec:exp_loss} will further elaborate on the effects of this modification with corresponding experimental results.

Formalization of the described process leads to the ultimate loss function employed by CoT-BERT, as illustrated in Equation~\ref{eq:loss_function_1} and \ref{eq:loss_function_2}. Here, $\boldsymbol r_i$ represents the template-denoised sentence embedding for $x_i$. Particularly, in this context, the standard InfoNCE Loss can be expressed as $\ell_i = - {\log} \frac{a_{ii^+}}{\sum_{j=1}^N a_{ij^+}}$. 

\begin{equation}
\begin{aligned}
&a_{ii^+} = {\exp}\Big(\frac{{\cos}(\boldsymbol r_i, \boldsymbol r_i^+)}{\tau}\Big)
 \qquad a_{ij^+} = {\exp}\Big(\frac{{\cos}(\boldsymbol r_i, \boldsymbol r_j^+)}{\tau}\Big)\\
&a_{ij^-} = {\exp}\Big(\frac{{\cos}(\boldsymbol r_i, \boldsymbol r_j^-)}{\tau}\Big)
 \qquad a_{i^+j^-} = {\exp}\Big(\frac{{\cos}(\boldsymbol r_i^+, \boldsymbol r_j^-)}{\tau}\Big)
\end{aligned}
\label{eq:loss_function_1}
\end{equation}
\begin{equation}
\text{ours:} \quad \mathlarger  \ell_i = - {\log} \frac{a_{ii^+}}{\sum_{j=1}^N \Big(a_{ij^+} + a_{ij^-} + a_{i^+j^-}\Big)}
\label{eq:loss_function_2}
\end{equation}

\subsection{Leveraging [PAD] Tokens for Template Denoising}
\label{sec:template_denoising}

Deriving sentence representations through prompts involves the concatenation of templates with input sentences, which are then fed into the encoder for a complete forward pass. However, due to the attention mechanism, the output vector of the [MASK] token is influenced by the presence of the template, potentially distorting its original semantics.  

To address this concern, PromptBERT suggests sending an empty template into BERT while maintaining position ids identical to those utilized when incorporating the input sentence. This process generates the template bias $\boldsymbol{\hat{h}}_i$. Ultimately, the embedding $\boldsymbol r_i$ used in the contrastive learning objective function is the difference between the sentence vector $\boldsymbol h_i$ and the template bias $\boldsymbol{\hat{h}}_i$.

From our perspective, a more effective approach than modeling input sentences with identical position ids is to populate [PAD] placeholders, matching the length of input sentences. This strategy naturally aligns position ids. Additionally, [PAD] placeholders can construct sentences devoid of significant meaning, thus serving to represent the inherent semantics of the template.

\begin{figure}[htbp]
\small
\centering
\includegraphics[width=\linewidth]{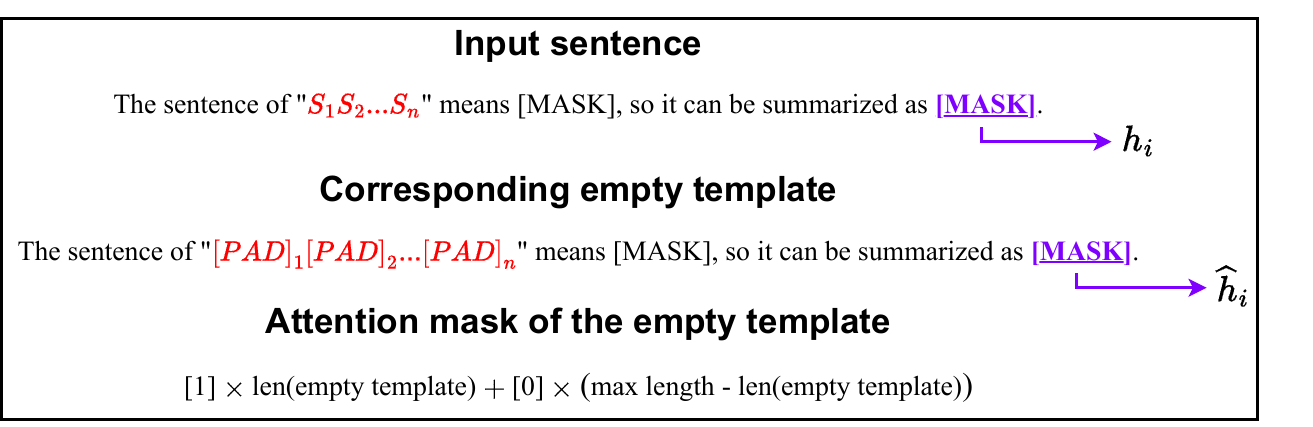}
\caption{Illustration of the template denoising method employed by CoT-BERT. In this depiction, we utilize the template for anchor sentences as an example, with analogous treatment applied to both positive and hard negative instances.}
\label{fig:pad_denoise}
\end{figure}

Figure~\ref{fig:pad_denoise} provides a detailed illustration of our template denoising method. In line with our two-stage representation derivation method, we extract the embedding corresponding to the final [MASK] token as the representation for the empty template. Since samples are processed in batches within the encoder, we pad each template to a predefined maximum length. Subsequently, we set the attention masks for the empty template to 1, while assigning 0 to the attention masks for the padded tail.

The comparison between our designed denoising strategy and the method proposed by PromptBERT, conducted on both BERT and RoBERTa, is discussed in subsection~\ref{sec:exp_denoise}.

\section{Experiment}
\label{sec:experiment}

\subsection{Setup}

In our experimental setup, we adhere to the established conventions by utilizing the SentEval \cite{SentEval-LREC-2018} toolkit for the evaluation of our model, CoT-BERT, across seven distinct English STS tasks \cite{STS12, STS13, STS14, STS15, STS16, STS-B, SICK-R}, with Spearman correlation as the metric.

Our training dataset is sourced from SimCSE \cite{SimCSE-EMNLP-2021}, comprising $10^6$ sentences randomly sampled from English Wikipedia. During the training phase, we save checkpoints based on our model's performance on the development set of STS-Benchmark. In the validation phase, we directly compute the cosine similarity between sentence embeddings without any additional regressors. Furthermore, akin to PromptBERT, we only perform template denoising during the training process. 

Our selection of baselines includes a diverse range of models, encompassing both non-BERT-based methods such as GloVe \cite{Glove-EMNLP-2014} and USE \cite{USE-EMNLP-2018}, as well as the latest advancements in unsupervised sentence representation built upon BERT architecture. Specifically, the models chosen for comparison include BERT-flow, BERT-whitening, IS-BERT \cite{IS-BERT-EMNLP-2020}, ConSERT, SimCSE, DCLR, ArcCSE, ESimCSE, DiffCSE, PCL, PromCSE, PromptBERT, ConPVP, RankEncoder, and RankCSE. This set of baselines serves to thoroughly validate the effectiveness of CoT-BERT from various perspectives.

\subsection{Main Results}

Table~\ref{tab:sts_result} provides a comprehensive overview of our primary experimental results across multiple STS datasets. When RoBERTa$_{\rm base}$ is employed as the encoder, CoT-BERT achieves an exceptional Spearman's correlation score of 80.62, setting a new record for the current best performance. In scenarios where BERT$_{\rm base}$ serves as the encoder, our model attains an average Spearman correlation of 79.40 on the seven STS tasks. While this value falls slightly below the existing SOTA results achieved by RankCSE and RankEncoder, it is imperative to note that CoT-BERT's architecture exclusively relies on BERT itself, whereas RankCSE and RankEncoder incorporate other text representation models such as SimCSE, SNCSE \cite{SNCSE-ICIC-2023}, or external database. Therefore, CoT-BERT's performance remains highly competitive and is obtained with a concise model configuration.

Moreover, compared to SimCSE, CoT-BERT outperforms it by 3.15\% and 4.05\% on BERT$_{\rm base}$ and RoBERTa$_{\rm base}$, respectively. In contrast to PromptBERT, which also leverages prompt-based techniques, CoT-BERT exhibits a 0.86\% improvement on BERT$_{\rm base}$ and a 1.47\% enhancement on RoBERTa$_{\rm base}$. These results collectively provide compelling evidence attesting to the effectiveness of CoT-BERT.
\begin{table*}[ht]
\caption{Performance of different models on STS tasks under \textbf{unsupervised} settings. Consistent with established research conventions, the table displays the Spearman's rank correlation between model predictions and human-annotated scores. Baseline results are sourced from original papers. The highest scores obtained by models using the same PLM for each dataset are highlighted in bold.}
\centering
\begin{adjustbox}{width=1.0\linewidth, center}
  \begin{tabular}{p{2.1cm}|l|cccccccc}
    \toprule
    \bf PLMs & \bf Methods  & \bf STS12 & \bf STS13 & \bf STS14 & \bf STS15 & \bf STS16 & \bf STS-B & \bf SICK-R & \bf Avg. \\
    \midrule
    \multirow{2}{*}{ Non-BERT} 
    & GloVe(avg.) & 55.14 & 70.66 & 59.73 & 68.25 & 63.66 & 58.02 & 53.76 & 61.32 \\ 
    & USE & 64.49 & 67.80 & 64.61 & 76.83 & 73.18 & 74.92 & 76.69 & 71.22 \\
    \midrule
    \multirow{16}{*}{BERT$_{\rm base}$} 
    & BERT-flow & 58.40 & 67.10 & 60.85 & 75.16 & 71.22 & 68.66 & 64.47 & 66.55 \\
    & BERT-whitening & 57.83 & 66.90 & 60.90 & 75.08 & 71.31 & 68.24 & 63.73 & 66.28 \\
    & IS-BERT & 56.77 & 69.24 & 61.21 & 75.23 & 70.16 & 69.21 & 64.25 & 66.58 \\
    & ConSERT & 64.64 & 78.49 & 69.07 & 79.72 & 75.95 & 73.97 & 67.31 & 72.74 \\
    & SimCSE & 68.40 & 82.41 & 74.38 & 80.91 & 78.56 & 76.85 & 72.23 & 76.25 \\
    & DCLR & 70.81 & 83.73 & 75.11 & 82.56 & 78.44 & 78.31 & 71.59 & 77.22 \\
    & ArcCSE & 72.08 & 84.27 & 76.25 & 82.32 & 79.54 & 79.92 & 72.39 & 78.11 \\
    & ESimCSE & 73.40 & 83.27 & 77.25 & 82.66 & 78.81 & 80.17 & 72.30 & 78.27 \\
    & DiffCSE & 72.28 & 84.43 & 76.47 & 83.90 & 80.54 & 80.59 & 71.23 & 78.49 \\
    & PCL & 72.84 & 83.81 & 76.52 & 83.06 & 79.32 & 80.01 & 73.38 & 78.42 \\
    & PromCSE & 73.03 & 85.18 & 76.70 & 84.19 & 79.69 & 80.62 & 70.00 & 78.49\\
    & PromptBERT & 71.56 & 84.58 & 76.98 & 84.47 & 80.60 & 81.60 & 69.87 & 78.54\\
    & ConPVP & 71.72 & 84.95 & 77.68 & 83.64 & 79.76 & 80.82 & 73.38 & 78.85\\
    & RankCSE$_{\rm listNet}$ & 74.38 & \bf 85.97 & 77.51 & 84.46 & \bf 81.31 & 81.46 & 75.26 & 80.05 \\
    & RankEncoder & \bf 74.88 & 85.59 & \bf 78.61 & 83.50 & 80.56 & 81.55 & \bf 75.78 & \bf 80.07\\
    & CoT-BERT & 72.56 & 85.53 & 77.91 & \bf 85.05 & 80.94 & \bf 82.40 & 71.41 & 79.40 \\
    \midrule
    \multirow{9}{*}{RoBERTa$_{\rm base}$} 
    & SimCSE & 70.16 & 81.77 & 73.24 & 81.36 & 80.65 & 80.22 & 68.56 & 76.57 \\
    & DCLR & 70.01 & 83.08 & 75.09 & 83.66 & 81.06 & 81.86 & 70.33 & 77.87 \\
    & DiffCSE & 70.05 & 83.43 & 75.49 & 82.81 & 82.12 & 82.38 & 71.19 & 78.21 \\
    & PCL & 71.13 & 82.38 & 75.40 & 83.07 & 81.98 & 81.63 & 69.72 & 77.90 \\
    & ESimCSE & 69.90 & 82.50 & 74.68 & 83.19 & 80.30 & 80.99 & 70.54 & 77.44\\
    & PromptRoBERTa & 73.94 & 84.74 & 77.28 & 84.99 & 81.74 & 81.88 & 69.50 & 79.15\\
    & ConPVP & 73.20 & 83.22 & 76.24 & 83.37 & 81.49 & 82.18 & \bf 74.59 & 79.18\\
    & RankCSE$_{\rm listNet}$ & 72.91 & \bf 85.72 & 76.94 & 84.52 & \bf 82.59 & \bf 83.46 & 71.94 & 79.73 \\
    & CoT-RoBERTa & \bf 75.43 & 85.47 & \bf 78.74 & \bf 85.64 & 82.21 & 83.40 & 73.46 & \bf 80.62 \\
    \bottomrule
\end{tabular}
\end{adjustbox}
\label{tab:sts_result}
\end{table*}

\subsection{Assessing our Two-stage Sentence Representation}
\label{sec:exp_cot}

To further ascertain the efficacy of CoT-BERT's two-stage sentence representation strategy and address the inquiries raised in subsection~\ref{sec:two_stage_representation}, we embark on an extensive series of experiments. For the manual templates provided in Table~\ref{tab:templates}, we define the initial and latter parts of the template as its \textit{prefix} and \textit{suffix}, respectively. To illustrate, for the anchor sentence template: ``The sentence of `[X]’ means [MASK], so it can be summarized as [MASK].'', we designate its \textit{prefix} as ``The sentence of `[X]’ means [MASK].'' and its \textit{suffix} as ``The sentence of `[X]’ can be summarized as [MASK].'' When we refer to ``\textit{prefix + suffix}'', it implies the use of our complete, original templates. The handling of positive instances and hard negative instances follows a similar pattern.

\textbf{Multi-stage vs. Single-stage}: Here, we explore whether CoT-BERT's two-stage sentence representation method surpasses the performance of each sub-stage when deployed independently. To this end, we conduct two sets of comparative experiments: \textit{prefix} vs. \textit{prefix + suffix} and \textit{suffix} vs. \textit{prefix + suffix}. If \textit{prefix + suffix} consistently performs better, it indicates that the integrative application of both stages effectively bolsters the model's performance.

\textbf{Progressive Relationship Between Sub-stages}: In this part, we replace CoT-BERT's original template \textit{prefix} with a non-sequitur (termed \textit{irrelevant prefix}) and examine whether the performance of \textit{irrelevant prefix + suffix} deteriorates compared to original \textit{prefix + suffix}, thereby proving the necessity of a logical connection between the two sub-stages. An example modification includes changing the anchor sentence template to ``Penguin is a flightless bird, and the sentence of `[X]' can be summarized as [MASK].'' The positive and hard negative instance templates are treated in the same manner.

Additionally, by reversing the order of the two sub-stages in CoT-BERT templates to \textit{suffix + prefix}, we seek to demonstrate the significance of the progressive reasoning process from comprehension to summarization. For instance, the anchor sentence template is converted to ``The sentence of `[X]' can be summarized as [MASK], so it means [MASK].'' The positive instance template and hard negative instance template are adjusted accordingly.

\textbf{Adaptivity in Multi-stage Reasoning}: Herein, we substitute the first [MASK] token in CoT-BERT templates with a static element (\textit{static prefix + suffix}) to observe performance disparities against the dynamic \textit{prefix + suffix} setup. This experiment aims to underscore the pivotal role of adaptive reasoning facilitated by the intermediate [MASK] token, notwithstanding that the derivation of sentence embeddings is from the final [MASK] token. For illustration, the anchor sentence template is changed to ``The sentence of `[X]' means something, so it can be summarized as [MASK].'', with corresponding adjustments made to the templates for positive and hard negative instances.

The results of these experiments are recorded in Table~\ref{tab:cot_result}, demonstrating that the \textit{prefix + suffix} configuration achieves the highest average Spearman's correlation score across seven STS benchmarks. These findings affirm the rationality and effectiveness of CoT-BERT's two-stage sentence representation strategy. Additionally, the specific settings and sequencing of the comprehension and summarization stages, together with the introduction of adaptive [MASK] tokens, adeptly emulate the essence of CoT reasoning.
\begin{table}[htbp]
\caption{Comparative analysis of CoT-BERT's two-stage sentence representation.} 
\renewcommand
\arraystretch{1.3}
\centering
\small
\setlength{\tabcolsep}{2pt}
\begin{tabular}{lc}
\hline
    & BERT$_\textrm{base}$ \\
\hline
only prefix & 78.91 \\
only suffix & 78.83 \\
irrelevant prefix + suffix & 78.44 \\
static prefix + suffix & 78.91 \\
suffix + prefix & 78.46 \\
prefix + suffix & \bf 79.40 \\
\hline
\end{tabular}
\label{tab:cot_result}
\end{table}

\subsection{Evaluating our Extended InfoNCE Loss}
\label{sec:exp_loss}

In subsection~\ref{sec:extended_infonce_loss}, we have expounded upon our proposed Extended InfoNCE Loss and its design concept. Departing from prior work, we introduce a comparison between the anchor sentence and negative samples in the loss function's denominator, as well as a comparison between the positive and negative instances, with the latter being proposed by CoT-BERT for the first time. In line with the symbol definitions in Equation~\ref{eq:loss_function_1}, we maintain the usage of $a_{i^+j^-}$ to denote the similarity computation between the positive and negative instances. 

Table~\ref{tab:loss_result} displays the performance disparities exhibited by CoT-BERT across seven STS tasks, both with and without the introduction of $a_{i^+j^-}$. The values in the table correspond to the average Spearman correlation. Remarkably, irrespective of whether BERT$_{\rm base}$ or RoBERTa$_{\rm base}$ serves as the PLM, our extended InfoNCE Loss consistently yields improvements in the model's performance.
\begin{table}[htbp]
\caption{Ablation experiments on the extended InfoNCE Loss for CoT-BERT.} 
\renewcommand
\arraystretch{1.2}
\centering
\small
\setlength{\tabcolsep}{2pt}
\begin{tabular}{lcc}
\hline
    & BERT$_\textrm{base}$ & RoBERTa$_\textrm{base}$ \\
\hline
CoT-BERT (without $a_{i^+j^-}$) & 79.11 & 80.30\\
CoT-BERT (with $a_{i^+j^-}$) & \bf 79.40 & \bf 80.62\\
\hline
\end{tabular}
\label{tab:loss_result}
\end{table}

To offer deeper insights into the underlying factors underpinning these results, we conduct alignment and uniformity analyses on three models: PromptBERT, CoT-BERT (with and without $a_{i^+j^-}$). This evaluation utilizes the STS-B test set, a corpus encompassing a total of 1,379 sentence pairs, each accompanied by a similarity score ranging from 0.0 to 5.0. 

For the computation of uniformity, the entire set of 1,379 sentence pairs is employed. In the case of alignment calculation, we filter for sentence pairs with similarity scores greater than 4.0. The results are detailed in Table~\ref{tab:align_uniform}.

Both alignment and uniformity serve as established metrics for evaluating the model’s semantic space quality. Given a data distribution $p_{\text{data}}$, alignment quantifies the expected distance between samples and their corresponding positive instances, defined as below:
\begin{equation}
    \ell_{\text{align}}\triangleq \underset{(x, x^+)\sim p_{\text{data}}}{\mathbb{E}} \Vert f(x) - f(x^+) \Vert^2
\end{equation}

Uniformity, on the other hand, reflects the overall evenness of the sentence vector space by calculating the average distance between embeddings of any two semantically unrelated texts:
\begin{equation}
    \ell_{\text{uniform}}\triangleq{\log} \underset{~~~x, y\stackrel{i.i.d.}{\sim} p_{\text{data}}}{\mathbb{E}}  e^{-2\Vert f(x)-f(y) \Vert^2}
\end{equation}

The experimental outcomes in Table~\ref{tab:align_uniform} provide empirical evidence in support of the intuitive explanations outlined in subsection~\ref{sec:extended_infonce_loss}. We posit that the incorporation of $a_{ij^-}$ introduces additional contextual references during the training process, thus enhancing the model's discriminative power among diverse samples. The inclusion of $a_{i^+j^-}$ further amplifies this effect by concurrently distinguishing negative instances from both anchor sentences and positive instances, thereby leading to an augmentation in the model's uniformity.
\begin{table}[htbp]
\caption{Results of alignment and uniformity calculations performed on models using the STS-B test set. Lower values indicate better performance. All three sets of experiments employed RoBERTa$_{\rm base}$ as the pre-trained model.} 
\renewcommand
\arraystretch{1.2}
\centering
\small
\setlength{\tabcolsep}{2pt}
\begin{tabular}{lcc}
\hline
PLM = RoBERTa$_\textrm{base}$ & Alignment & Uniformity \\
\hline
PromptBERT & 0.0957 & - 1.2033\\
CoT-BERT (without $a_{i^+j^-}$) & 0.1089 & - 1.3852\\
CoT-BERT (with $a_{i^+j^-}$) & 0.1278 & - 1.5492\\
\hline
\end{tabular}
\label{tab:align_uniform}
\end{table}

Meanwhile, a diminution in the alignment metric is observed. We attribute this phenomenon to two potential factors. Firstly, while computing alignment with the STS-B test set, we categorize sentence pairs with a similarity exceeding 4.0 as positive examples, a threshold selection that may introduce bias. Additionally, when modifying the objective function to impose more stringent requirements on the model, there is no corresponding increase in the number of model iterations. Consequently, within a constrained number of update steps, the model may struggle to optimally adjust its alignment. 

\subsection{Assessing our Template Denoising Strategy}
\label{sec:exp_denoise}

We also embark on an ablation study regarding the template denoising method of CoT-BERT. To distinguish between the two denoising strategies under examination, we refer to the technique introduced by PromptBERT as ``position denoise'' and our proposed approach as ``[PAD] denoise.''

The primary difference between these two lies in the fact that ``[PAD] denoise'' does not deliberately adjust the values of position ids but rather models an empty template by injecting [PAD] placeholders of identical length as the input sentence, accompanied by corresponding adaptations to the attention masks.

Our evaluation of different denoising methods on the seven STS tasks is presented in Table~\ref{tab:denoise_result}. As in our prior assessments, we continue to report the model's average Spearman correlation. 
\begin{table}[htbp]
\caption{Ablation study of CoT-BERT's template denoising strategy.} 
\renewcommand
\arraystretch{1.2}
\centering
\small
\setlength{\tabcolsep}{2pt}
\begin{tabular}{lcc}
\hline
& BERT$_\textrm{base}$ & RoBERTa$_\textrm{base}$ \\
\hline
CoT-BERT (without denoise) & 78.69 & 79.87 \\
CoT-BERT (position denoise) & 78.89 & 79.95\\
CoT-BERT ([PAD] denoise) & \bf 79.40 & \bf 80.62\\
\hline
\end{tabular}
\label{tab:denoise_result}
\end{table}

The empirical findings compellingly demonstrate the advantages conferred by the introduction of denoising encodings within the contrastive learning loss function. Moreover, ``[PAD] denoise'' notably outperforms its ``position denoise'' counterpart.

One possible explanation for the effectiveness of template denoising is that sentence embeddings derived from PLMs may contain some general information, such as syntax and sentence structure. During the contrastive learning process, subtracting that kind of information helps highlight the distinctions between input sentences, thereby fostering more precise clustering outcomes.

\section{Discussion}

\subsection{Distribution of Predicted Values}

\begin{figure*}[htbp]
\centering
\includegraphics[width=1.0\linewidth]{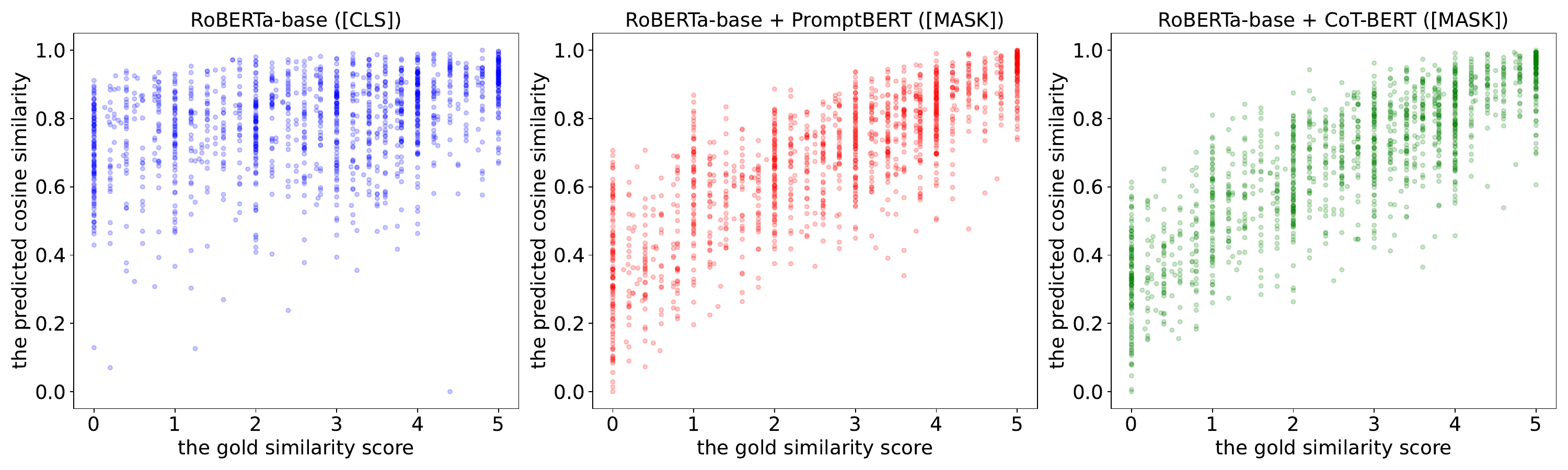}
\caption{Correlation diagram between the true similarity scores and model-predicted cosine similarity on the STS-B test set. The vertical axis has been normalized for clarity, and the methods employed for deriving sentence embeddings ([CLS] or [MASK]) are explicitly indicated for reference.}
\label{fig:distribution}
\end{figure*}

We present a visualization of the predicted value distribution generated by CoT-BERT using the STS-B test set in Figure~\ref{fig:distribution}. It should be noted that the model remains unexposed to these data during its training phase. Therefore, the model's performance on this dataset can serve as a reliable indicator of its overall efficacy.

As illustrated in Figure~\ref{fig:distribution}, the initial RoBERTa checkpoint demonstrates limited discriminative ability for sentence pairs with varying degrees of similarity. It tends to yield higher predicted values across the board. In contrast, both PromptBERT and CoT-BERT display a clear upward trend in predicted values as the similarity between sentences increases.

Furthermore, CoT-BERT distinctly outperforms PromptBERT, especially in handling samples with annotated similarity scores ranging from 0 to 2. CoT-BERT's predictions within this range are considerably more concentrated, indicating enhanced precision. Additionally, it is noteworthy that for some samples with a true similarity score of 1, PromptBERT's predicted values are even higher than those for most samples with a true similarity score of 2, while CoT-BERT does not exhibit such a pattern.

\subsection{Introducing More Stages}

In subsection~\ref{sec:two_stage_representation}, we have elucidated the underlying design principles behind CoT-BERT's two-stage sentence representation method. Furthermore, in subsection~\ref{sec:exp_cot}, we have empirically demonstrated the superiority of this approach over using either sub-stage in isolation. Naturally, this leads to a pertinent question: if we further divide the template and introduce more sub-stages, will the model's performance continue to improve?

Regrettably, due to constraints on computational resources, we are currently unable to conduct experiments in this regard. Nevertheless, we recognize several crucial considerations that warrant careful attention in such investigations.

Firstly, increasing the number of stages within the template inherently augments the complexity of the corresponding prompt. This heightened complexity demands substantial effort in devising and selecting the most suitable prompts. Additionally, even if a template performs well on a specific PLM, its adaptability to other PLMs remains uncertain. Besides, as the template's length expands, the weight of the input sentence [X] within the prompt gradually diminishes, and its distance from the final [MASK] token increases. This could potentially result in the model inadequately capturing the semantics of [X]. Moreover, it's worth mentioning that certain concise short sentences in natural language may not lend themselves well to being segmented into multiple stages. Lastly, the presence of templates compresses the maximum input length that a PLM can accommodate, thereby affecting the model's capacity to handle longer texts. We leave the exploration of these aspects to future work.

\section{Conclusion}

In this study, we propose CoT-BERT, a pioneering strategy for sentence representation computation. To the best of our knowledge, CoT-BERT is the first work that combines the Chain-of-Thought (CoT) concept with text representation tasks. Furthermore, we improve the distribution of sentence embeddings within the BERT semantic space by introducing an extended InfoNCE Loss. Additionally, we devise a more efficacious template denoising method to mitigate the impact of prompt-induced biases on sentence semantics.

Experimental findings across seven Semantic Textual Similarity tasks unequivocally affirm the outstanding efficacy of CoT-BERT. It surpasses a spectrum of formidable baselines and achieves state-of-the-art performance without relying on additional text representation models or external databases. Comprehensive ablation experiments demonstrate that CoT-BERT's two-stage sentence representation, extended InfoNCE loss, and refined template denoising methods collectively contribute to the enhancements in its overall performance.


\end{document}